# 基于邻域区间扰动融合的无监督特征选择算法框架

吕晓林[1],杜 亮[1,2],周 芃[3],吴 鹏[1,2]

(山西大学 1.计算机与信息技术学院;2.大数据科学与产业研究院,山西 太原 030006;
3.安徽大学 计算机科学与技术学院,安徽 合肥 230601)

**摘 要**:特征选择技术是数据降维的一种关键技术,由于采集到的数据样本标签信息缺失,无监督特征选择受到了更多人的关注。现有的无监督特征选择算法普适性及稳定性很低,受数据集结构的影响很大,因此很多研究者一直热衷于提高算法的稳定性。该文尝试从数据集的预处理出发,采用区间的方式来对数据集进行近似,得到与数据集相关联的几个数据集,通过实验验证新的区间数据集的优劣性,并思考从全局的角度对数据集进行处理,进一步提出了一种新的模型——基于邻域区间扰动融合的无监督特征选择算法框架(Unsupervised feature selection algorithm framework based on neighborhood interval disturbance fusion,NIDF)。该模型可实现对特征的最终得分和近似数据区间的联合学习,通过与原始无监督特征选择方法以及现有的几种特征选择框架的对比,体现出该文提出的模型的优越性。

# Unsupervised feature selection algorithm framework based on neighborhood interval disturbance fusion

Lv Xiaolin[1],Du Liang[1,2],Zhou Peng[3],Wu Peng[1,2]

(1.College of Computer and Information Technology;2. Institute of Big Data Science and Industry,
Shanxi University,Taiyuan 030006,China;
3.College of Computer Science and Technology,Anhui University,Hefei 230601,China)

**Abstract**:Feature selection technology is a key technology of data dimensionality reduction. Because of the lack of label information of collected data samples,unsupervised feature selection has attracted more attention. The universality and stability of many unsupervised feature selection algorithms are

**收稿日期**:2021-05-08 **修回日期**:2021-07-07

**基金项目**:国家自然科学基金(61502289;61806003);山西省重点研发项目(201803D31199);山西省自然科学基金(201801D221163;201801D221173)

**作者简介**:吕晓林(1995-),女,硕士生,主要研究方向:机器学习、数据挖掘,E-mail:969103861@qq.com;通讯作者:杜亮(1985-),男,博士,副教授,主要研究方向:数据挖掘、机器学习、大数据分析,E-mail:152009746@qq.com。

very low and greatly affected by the dataset structure. For this reason, many researchers have been keen to improve the stability of the algorithm. This paper attempts to preprocess the data set and use an interval method to approximate the data set, experimentally verifying the advantages and disadvantages of the new interval data set. This paper deals with these data sets from the global perspective and proposes a new algorithm—unsupervised feature selection algorithm based on neighborhood interval disturbance fusion(NIDF). This method can realize the joint learning of the final score of the feature and the approximate data interval. By comparing with the original unsupervised feature selection methods and several existing feature selection frameworks, the superiority of the proposed model is verified.

高维大数据在许多领域随处可见，科技的进步更是加快了大数据的产生，每天都会有数亿的数据产生，以文本、图像、音频、视频等形式存在，覆盖于各个领域。面对如此庞大的数据，从中选择出合适的信息对学习工作进行指导变得极为困难。在这种境况下，特征选择技术显得尤为重要。特征选择技术的主要目的是在一个特定的评估标准下，从原始的高维特征中选择出最重要的特征子集，然后利用选择出的特征子集结合一些有效的算法去完成数据聚类、分类等任务。

根据数据样本是否含有标签信息，特征选择算法可分为有监督特征选择[1,2]、半监督特征选择[3-5]和无监督特征选择[6-8]3类。有监督和半监督特征选择通常会用到样本的标签信息，通过特征和标签信息之间的相关性来评定特征的重要性。现实中采集到的数据很少有标签信息，并且标记标签信息的代价很昂贵，在大规模数据中更是难以实现，因此，研究无监督场景下的特征选择更具有实用价值。

无监督特征选择方法可分为3类：过滤式方法[9]、包装式方法[10]和嵌入式方法[11-13]。过滤式方法的主要思路是对每一维的特征打分，即给每一维的特征赋予权重，权重代表该维特征的重要性，然后依据权重排序。过滤式方法是独立于学习算法的，它的计算量很低，它的性能也有所欠缺。包装式方法是将子集的选择看作是一个搜索寻优的问题，首先生成不同的组合，对组合进行评价，然后再与其他的组合进行比较。包装式方法是与特定的学习算法相联系的，虽然有较好的性能，但是它的计算量很大，不适用于对大规模数据的处理。嵌入式方法的思路是在模型既定的情况下得出对提高模型准确性最好的属性，即在确定模型的过程中，选出对模型有重要意义的属性。具体而言，嵌入式方法是将特征选择的学习过程嵌入进模型中，故嵌入式方法能获得一个较好的性能，但其计算量很大，不具备通用性。

## 1　相关工作

近年来，研究者基于无监督特征选择方法已经做了大量的工作，其中最具有代表性的是拉普拉斯特征选择算法(Laplacian score for feature selection, LapScore)[14]和多类簇特征选择算法(Unsupervised feature selection for multi-cluster data, MCFS)[15]。LapScore是一种基于过滤式的无监督特征选择方法，是由何晓飞于2005年提出的，LapScore主要是基于拉普拉斯特征映射和局部保留投影，它的主要思想是距离近的两个样本点更可能属于同一类。因此，LapScore认为，数据的局部结构比全局结构更重要。LapScore是通过特征保留数据局部结构的能力来评估其重要性的。MCFS是一种基于嵌入式的无监督特征选择方法，是由蔡登于2010年提出的，其主要思想是通过谱分析展开数据流形，然后利用L1谱分析回归来决定特征的重要性。这两种方法都是从改造模型的角度出发进行特征选择，并且都没有考虑特征的冗余性，所选出的特征具有很大的冗余性。对此，Wang等[16]于2015年提出了一种基于全局冗余最小化的特征选择框架(Feature selection via global redundancy minimization, GRM)。进一步地，Nie等[17]于2019年提出了一种基于全局冗余最小化的自动加权特征选择框架(General frame-work for auto-weighted feature selection via global redun-dancy minimization，AGRM）。这两个框架都能够选择出不冗余的特征，既适用于有监督特征选择方法，也适用于无监督特征选择方法。

另外，在以往的无监督特征选择工作中发现，对不同数据集进行降维，选择特征的能力是不同的，甚至有很大差异。数据的准确与否会对所选择的特征产生很大影响，提出的模型抗干扰能力差，性能极易受到个别异常点的破坏。为此，研究者们从不同的维度提出了一系列的解决办法，包括自步学习、鲁棒学习等。

（1）自步学习。

自步学习是近年来被提出的一种学习策略，它从简单到复杂逐步增加训练实例，可以典型地降低局部最优的风险。基于聚类任务现存的一些问题，如聚类结果很容易陷入局部最优、聚类结果容易受到少量异常值的影响、聚类结果对初始参数非常敏感等，很多研究者发现，通过加入自步学习框架可以大大缓解此类问题。具体地，Yu 等[18]于2020年提出了一种基于自步学习的K-means聚类算法。其核心思想是通过自步学习方法来模拟人类学习知识的过程，即从易到难地学习知识。该方法首先使用一个自步正则化因子来选择样本的一个特定子集加入训练，然后自动调整自步学习的参数逐步地增加样本的数量和难度，从而逐渐提高聚类模型的性能和泛化能力。

（2）鲁棒学习。

为了克服异常点对模型的影响，增强算法的稳定性，研究者们针对聚类任务提出了许多鲁棒的算法[19,20]。这些方法大致可归为两类：一类是基于惩罚正则化的方法，另一类是基于裁剪函数的方法。Georgogiannis 于2016年通过借鉴回归中离群点检测的思想，提出了一个经典二次K-均值算法的变量鲁棒性和一致性的理论分析——鲁棒 K-means[21]。在这项工作中，Georgogiannis 发现，一个数据集中的两个离群值足以分解这个聚类过程。然而，如果关注的是“结构良好的”数据集，那么尽管有离群值，鲁棒 K-means 最终还是可以恢复底层的集群结构。

可以看出，针对数据集中的离群点，为了降低其对模型性能的影响，研究者们都是努力地提升模型的稳定性，不可否认的是，这种做法确实能缓解离群点对模型整体性能的破坏，但是，正如 Georgogiannis 在鲁棒 K-means 的工作中所提出的，对于“结构良好的”数据集，鲁棒 K-means 可以克服离群值的影响，恢复数据的底层集群结构，而对于某些数据集，这个数据集中的两个离群值足以破坏整个聚类过程，这足以说明对于一个“非结构良好的”数据集，离群值的破坏力很大。

为此，本文从两方面来提高聚类的准确性。一方面，采用区间的方式对数据进行近似，相较于直接使用某个样本自身的信息，本文采用其邻近的几个样本来刻画这个样本，这样可以有效地降低离群点的影响。另一方面，基于上述产生的多种数据表示，提出了一种新的模型——基于邻域区间扰动融合的无监督特征选择算法框架（Unsupervised feature selection algorithm framework based on neighborhood interval disturbance fusion，NIDF）。此模型可实现对特征的最终得分和近似数据区间的联合学习，最终达到一个不错的聚类效果。

## 2 研究方法

### 2.1 方法的提出

在这项工作中，本文首先对数据进行了区间近似，具体地，给定一个数据集 $X \in \mathbf{R}^{n\times d}$，本文从两个层面对这个数据集中的数据进行近似。一方面，从样本层面出发，采用K近邻的方式找到每个样本的 $k$ 个邻居，然后采用方差浮动的方式将样本近似到一个区间上，得到近似数据集 $X_{\text{low}}$ 和 $X_{\text{up}}$；另一方面，从特征层面出发，同样采用K近邻的方式找到每个特征的 $k$ 个邻居，将特征近似到一个区间上，得到近似数据集 $F_{\text{low}}$ 和 $F_{\text{up}}$，这样，就将一个原始数据集 $X$ 扩展成了4个近似数据集，这里采用 $\{X_i\}_{i=1}^{4}$ 来表示，$\{X_i\}_{i=1}^{4}$ 分别对应于 $X_{\text{low}}$、$X_{\text{up}}$、$F_{\text{low}}$ 和 $F_{\text{up}}$。其次，本文采用经典的无监督特征选择方法对新的区间近似数据集 $\{X_i\}_{i=1}^{4}$ 进行特征选择，并分别得到一个特征分数 $\{s_i\}_{i=1}^{4}$。最后，基于此，本文给出了一种新的模型——基于邻域区间扰动融合的无监督特征选择算法框架 NIDF。通过 NIDF 模型联合学习特征选择和近似的数据区间，得到最终的特征分数。具体的算法过程示意图如图1所示。

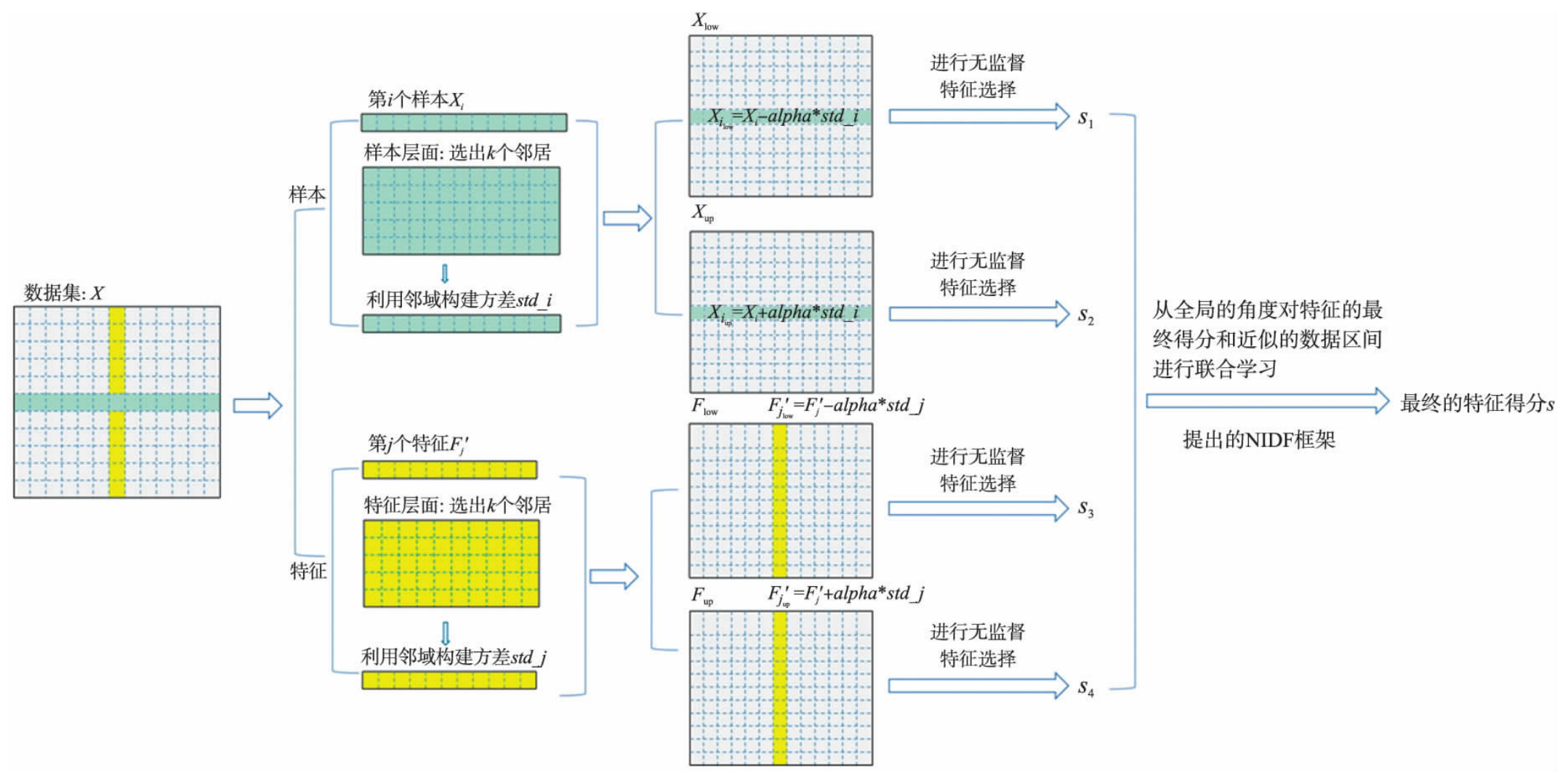

**图 1　算法过程示意图**

NIDF 模型如下

$$\min_{\lambda,z,w}\lambda^{2}\sum_{i=1}^{4}w_i^2 z^{\mathrm{T}}A_i z-\lambda\sum_{i=1}^{4}w_i z^{\mathrm{T}}s_i \qquad (1)$$

$$\text{s.t.}\quad z^{\mathrm{T}}1=1,z\geqslant 0\quad w^{\mathrm{T}}1=1,w\geqslant 0$$

式中: $A_i\in\mathbf{R}^{d\times d}$是一个冗余矩阵,用来刻画第 $i$ 个近似数据集上所有特征间的冗余性; $s_i\in\mathbf{R}^{d\times 1}$代表的是对第 $i$ 个近似数据集进行无监督特征选择后得到的特征分数; $w_i$ 代表的是第 $i$ 个近似数据集的权重; $\lambda$ 是一个自动加权的参数,用来平衡第一项和第二项; $z\in\mathbf{R}^{d\times 1}$是利用式(1)联合学习特征选择和近似的数据区间后得出的最终特征分数。

## 2.2　方法的优化

式(1)可通过迭代地更新 $\lambda$、$z$、$w$ 来求解,详细过程如下。

(1)固定 $z$ 和 $w$,更新 $\lambda$。则式(1)可等价于求解以下问题

$$\min_{\lambda}\lambda^{2}z^{\mathrm{T}}Az-\lambda z^{\mathrm{T}}s \qquad (2)$$

式中: $A\in\mathbf{R}^{d\times d}$, $A=\sum_{i=1}^{4}w_i^2A_i$, $s=\sum_{i=1}^{4}w_is_i$。

对于式(2),通过求其关于 $\lambda$ 的导数,并令其等于 0,则可求出

$$\lambda=\frac{z^{\mathrm{T}}s}{2z^{\mathrm{T}}Az} \qquad (3)$$

(2)固定 $\lambda$ 和 $w$,更新 $z$。则式(1)可等价于求解以下问题

$$\min_{z^{\mathrm{T}}1=1,z\geqslant 0}\lambda z^{\mathrm{T}}Az-z^{\mathrm{T}}s \qquad (4)$$

式中: $A\in\mathbf{R}^{d\times d}$, $A=\sum_{i=1}^{4}w_i^2A_i$, $s=\sum_{i=1}^{4}w_is_i$。

很明显,上述问题是一个带有线性约束的凸二次规划问题,该问题可用现有的优化工具轻松解决。

(3)固定 $\lambda$ 和 $z$,更新 $w$。则式(1)可等价于求解以下问题

$$\min_{w^{\mathrm{T}}1=1,w\geqslant 0}\lambda\sum_{i=1}^{4}w_i^2z^{\mathrm{T}}A_iz-\sum_{i=1}^{4}w_iz^{\mathrm{T}}s_i \qquad (5)$$

这里,引入矩阵 $H\in\mathbf{R}^{d\times d}$, $f\in\mathbf{R}^{d\times 1}$,其中 $H$ 是一个对角矩阵,其主对角线元素为 $H_{ii}=z^{\mathrm{T}}A_iz$,矩阵 $f$ 中的元素为 $f_i=z^{\mathrm{T}}s_i$。则式(5)可被重写为

$$\min_{w^{\mathrm{T}}1=1,w\geqslant 0}\lambda w^{\mathrm{T}}Hw-f^{\mathrm{T}}w \qquad (6)$$

同样,这是一个带有线性约束的凸二次规划问题,可用现有的优化工具解决。

总体来说,式(1)的解法如算法 1 所示。另外,整个 NIDF 模型的过程总结在算法 2 中。

**算法 1**　式(1)的优化解法

输入: $\{A_i\}_{i=1}^{4}$, $\{s_i\}_{i=1}^{4}$;
初始化: $z$, $w$;
重复:
1. 更新 $\lambda$ 通过式(3);
2. 更新 $z$ 通过解决式(4);
3. 更新 $w$ 通过解决式(6);

直到 $\lambda$ 收敛。
输出: $z$

**算法 2**　NIDF 模型的过程

输入: 数据集 $X$,样本标签 $Y$,所选特征数 $m$;
步骤:
1. 对数据集 $X$ 进行区间近似,得到其近似数据集$\{X_i\}_{i=1}^{4}$;
2. 采用经典的无监督算法对新的区间近似数据集$\{X_i\}_{i=1}^{4}$进行特征选择,并分别得到一个特征分数$\{s_i\}_{i=1}^{4}$;
3. 利用算法 1 求解 NIDF 模型,实现对特征选择和近似的数据区间联合学习,得出最终的特征分数 $z$;
4. 对特征分数 $z$ 进行降序排序,并选出其前 $m$ 个特征。

输出: 选出前 $m$ 个特征。

# 3 实验

本节通过在8个公开的数据集上进行实验，验证本文所提模型的有效性。

## 3.1 数据集

在实验中，本文使用了多种数据集，包括2个文本数据集，4个图像数据集，1个生物数据集，1个其他数据集，这些数据集都是进行特征选择的常用数据集，数据集的大小如表1所示。

**表1 数据集的大小**

| 数据集 | 样本数 | 特征数 | 类数 |
|---|---|---|---|
| PIE | 1428 | 1024 | 68 |
| 20NG | 3970 | 1000 | 4 |
| BBCNEWS | 737 | 1000 | 5 |
| AR | 840 | 768 | 120 |
| ORL | 400 | 4096 | 40 |
| RELATHE | 1427 | 4322 | 2 |
| CARCINOMAS | 174 | 9182 | 11 |
| USPS | 9298 | 256 | 10 |

## 3.2 实验对比算法

提出的模型作为一个后处理过程，可用于无监督特征选择方法中，这里使用了3个现有的无监督特征选择方法：拉普拉斯算法（LapScore）、多类簇特征选择算法（MCFS）和基于图结构的Kullback-Leibler（KL）散度最小化算法（Gragh-based Kullback-Leibler divergence minimization for unsupervised selection，KLMFS）。

另外，GRM和AGRM框架也可用于无监督特征选择中的后处理过程，此处将这两个框架应用于已有的无监督特征选择方法，可以得出以下几组对比算法。

（1）第一组。

LapScore[14]：基于原始数据集，利用拉普拉斯特征选择算法进行特征选择。

LapScore_GRM[16]：基于原始数据集，将GRM框架应用于拉普拉斯特征选择算法中进行特征选择。

LapScore_AGRM[17]：基于原始数据集，将AGRM框架应用于拉普拉斯特征选择算法中进行特征选择。

LapScore_NIDF：新提出的方法，基于构建的区间近似数据集，将NIDF框架应用于拉普拉斯特征选择算法中进行特征选择。

（2）第二组。

MCFS[15]：基于原始数据集，利用多类簇特征选择算法进行特征选择。

MCFS_GRM：基于原始数据集，将GRM框架应用于多类簇特征选择算法中进行特征选择。

MCFS_AGRM：基于原始数据集，将AGRM框架应用于多类簇特征选择算法中进行特征选择。

MCFS_NIDF：新提出的方法，基于构建的区间近似数据集，将NIDF框架应用于多类簇特征选择算法中进行特征选择。

（3）第三组。

KLMFS[22]：基于原始数据集，利用KL散度最小化算法进行特征选择。

KLMFS_GRM：基于原始数据集，将GRM框架应用于KL散度最小化算法中进行特征选择。

KLMFS_AGRM：基于原始数据集，将AGRM框架应用于KL散度最小化算法中进行特征选择。

KLMFS_NIDF：新提出的方法，基于构建的区间近似数据集，将NIDF框架应用于KL散度最小化算法中进行特征选择。

## 3.3 评价指标

在实验中，使用了聚类方法常用的两个评价指标来评估方法的性能，即聚类准确性（Accuracy，ACC）和归一化互信息（Normalized mutual information，NMI）。这两个指标的值越大，表示聚类性能越好。

聚类准确性：聚类准确性主要是用来刻画所聚的类和样本原始类之间的一对一关系。给定一个样本点 $x_i$，$p_i$ 和 $q_i$ 分别用来表示聚类结果和样本的真实标签，则 $ACC$ 为

$$ACC=\frac{1}{n}\sum_{i=1}^{n}\delta(q_i,map(p_i))$$

式中：$n$ 是样本数，$\delta(x,y)$ 是一个这样的函数，若 $x=y$，$\delta(x,y)$ 的值为1，否则为0。$map(\cdot)$ 是一个置换函数，将每一个簇索引映射到一个真实的类标签中。

归一化互信息：归一化互信息主要是用来刻画聚类的质量。记 $C$ 是真实类标签的集合，$C'$ 是通过聚类算法计算的类集合，则它们的互信息 $MI(C,C')$ 为

$$MI(C,C')=\sum_{c_i\in C,c_j'\in C'}p(c_i,c_j')\log\frac{p(c_i,c_j')}{p(c_i)p(c_j')}$$

式中：$p(c_i)$ 和 $p(c_j)$ 分别是从数据集中任意选定的一个样本点属于类 $c_i$ 和 $c_j$ 的概率，$p(c_i, c_j)$ 是这个数据点同时属于这两个类的概率。实验中，使用的归一化互信息 $NMI$ 为

$$NMI(C, C') = \frac{MI(C, C')}{\max(H(C), H(C'))}$$

式中：$H(C)$ 和 $H(C')$ 分别是 $C$ 和 $C'$ 的熵。

### 3.4　实验结果分析

本文进行了多组实验，以验证所提模型的有效性。

首先，将数据集 $X$ 进行区间近似，得到其近似数据集 $\{X_i\}_{i=1}^{4}$，接着，在原始数据集 $X$ 和近似数据集 $\{X_i\}_{i=1}^{4}$ 上进行无监督特征选择。这里以 PIE 数据集中的第一个样本图像进行 LapScore 特征选择为例。实验得出 LapScore 在原始数据集 $X$ 上的 $ACC$ 值为 0.556 0，在近似数据集 $\{X_i\}_{i=1}^{4}$ 上的 $ACC$ 值分别为 0.335 0、0.418 6、0.590 2 以及 0.433 3，说明对数据集做区间近似在某些情况下可以增强特征选择方法的性能，进一步说明区间近似数据集对特征选择性能的积极作用，直观的特征选择效果如图 2 所示。

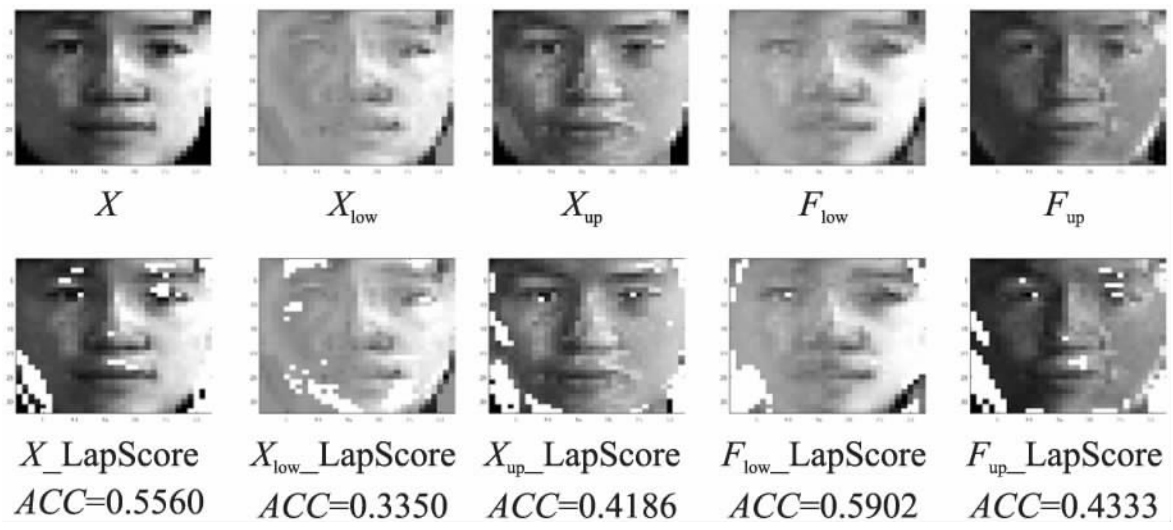

**图 2　原始数据集和区间近似数据集上 LapScore 的特征选择效果**

其次，针对区间近似数据集对特征选择的不稳定影响，包括积极的和消极的，本文联合学习区间近似数据集和特征选择——NIDF 模型。这里同样以 PIE 数据集中的第一个样本图像进行 LapScore 特征选择为例。在这组实验中，本文以原始的 LapScore 特征选择方法、经 GRM 和 AGRM 处理过的 LapScore_GRM 和 LapScore_AGRM 方法作为对比，实验得出，经本文提出的 NIDF 处理后的 LapScore_NIDF 能一定程度上提高特征选择的效果。直观的特征选择效果如图 3 所示，从图中可以看出，经 NIDF 处理后的 LapScore_NIDF 选择出的特征更集中，更具有代表性。

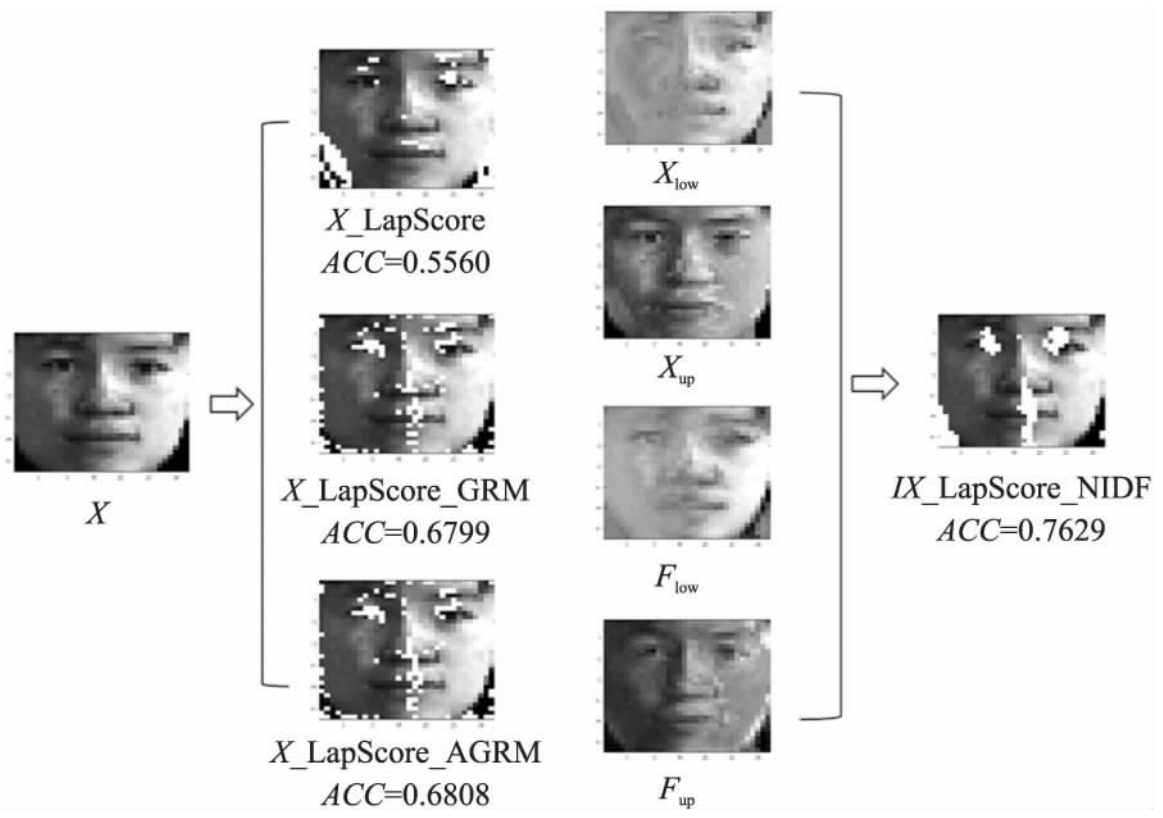

**图 3　3 种框架基于 LapScore 的特征选择效果**

最后，通过在大批量的数据集上分别进行 3 组对比算法的特征选择能力测试，可以看出本文提出的模型能一定程度上提高无监督特征选择方法的性能，聚类结果分别如表 2 和表 3 所示，其中最好的结果加粗表示，次好的结果加下划线表示。另外，3 组算法在不同数据集上 $ACC$ 值的直观展示如图 4 所示。

**表 2　不同数据集上的聚类 *ACC***

| 数据集 | LapScore | LapScore_GRM | LapScore_AGRM | LapScore_NIDF | MCFS | MCFS_GRM | MCFS_AGRM | MCFS_NIDF | KLMFS | KLMFS_GRM | KLMFS_AGRM | KLMFS_NIDF |
|---|---|---|---|---|---|---|---|---|---|---|---|---|
| PIE | 0.5349 | 0.6823 | 0.6915 | **0.7501** | 0.3653 | **0.6221** | 0.5720 | 0.5685 | 0.5857 | 0.6501 | 0.6314 | **0.7043** |
| 20NG | 0.2602 | 0.2614 | 0.2615 | **0.3302** | 0.2944 | 0.2690 | 0.2932 | **0.4016** | 0.2795 | 0.2644 | 0.2654 | **0.3766** |
| BBCNEWS | 0.5131 | 0.4556 | 0.5097 | **0.5246** | 0.5157 | 0.4835 | 0.4982 | **0.5497** | 0.5516 | 0.4810 | 0.5129 | **0.5550** |
| AR | 0.3015 | 0.3381 | 0.3529 | **0.3582** | 0.3228 | 0.3442 | 0.3399 | **0.3493** | 0.3556 | **0.3727** | 0.3654 | 0.3610 |
| ORL | 0.3887 | 0.4831 | **0.4911** | 0.4748 | 0.4630 | 0.4727 | **0.4735** | 0.4716 | 0.3542 | **0.4625** | 0.4548 | 0.4520 |
| RELATHE | 0.5395 | 0.5423 | 0.5377 | **0.5548** | 0.5446 | **0.5455** | 0.5451 | 0.5451 | 0.5461 | 0.5459 | 0.5460 | **0.5514** |
| CARCINOMAS | 0.5998 | 0.5186 | **0.6812** | 0.6425 | 0.5182 | 0.4322 | 0.5199 | **0.5519** | 0.6174 | 0.5881 | 0.6305 | **0.6744** |
| USPS | 0.4985 | 0.5690 | 0.5805 | **0.6373** | 0.4369 | 0.5736 | **0.5821** | 0.5514 | 0.5893 | 0.5804 | 0.5749 | **0.6113** |
| AVERAGE | 0.4545 | 0.4813 | 0.5133 | **0.5341** | 0.4326 | 0.4679 | 0.4780 | **0.4986** | 0.4849 | 0.4931 | 0.4977 | **0.5358** |

**表 3　不同数据集上的聚类 *NMI***

| 数据集 | LapScore | LapScore_GRM | LapScore_AGRM | LapScore_NIDF | MCFS | MCFS_GRM | MCFS_AGRM | MCFS_NIDF | KLMFS | KLMFS_GRM | KLMFS_AGRM | KLMFS_NIDF |
|---|---|---|---|---|---|---|---|---|---|---|---|---|
| PIE | 0.8157 | 0.8833 | 0.8896 | **0.9251** | 0.6875 | **0.8480** | 0.8218 | 0.8215 | 0.8294 | 0.8675 | 0.8576 | **0.9015** |
| 20NG | 0.0392 | 0.0406 | 0.0416 | **0.1252** | 0.0797 | 0.0346 | 0.0705 | **0.2235** | 0.0697 | 0.0426 | 0.0454 | **0.1857** |
| BBCNEWS | 0.3731 | 0.2410 | 0.3295 | **0.3794** | 0.3741 | 0.3386 | 0.3468 | **0.3924** | 0.3705 | 0.2824 | 0.3175 | **0.4205** |
| AR | 0.6651 | 0.6777 | 0.6873 | **0.6934** | 0.6709 | 0.6788 | 0.6743 | **0.6808** | 0.6936 | **0.6999** | 0.6980 | 0.6939 |
| ORL | 0.6222 | 0.6967 | **0.7056** | 0.6936 | 0.6852 | 0.6884 | **0.6896** | 0.6888 | 0.6082 | **0.6862** | 0.6809 | 0.6793 |
| RELATHE | 0.0082 | 0.0212 | 0.0225 | **0.0284** | **0.0077** | 0.0022 | 0.0058 | 0.0076 | 0.0043 | 0.0042 | 0.0042 | **0.0135** |
| CARCINOMAS | 0.6505 | 0.5690 | **0.7266** | 0.6844 | 0.5111 | 0.3952 | 0.5120 | **0.5438** | 0.6675 | 0.6246 | 0.6842 | **0.7175** |
| USPS | 0.4721 | 0.5053 | 0.5178 | **0.5634** | 0.3883 | 0.5245 | **0.5418** | 0.4810 | **0.5508** | 0.5274 | 0.5211 | 0.5473 |
| AVERAGE | 0.4558 | 0.4544 | 0.4901 | **0.5116** | 0.4256 | 0.4388 | 0.4578 | **0.4799** | 0.4743 | 0.4669 | 0.4761 | **0.5199** |

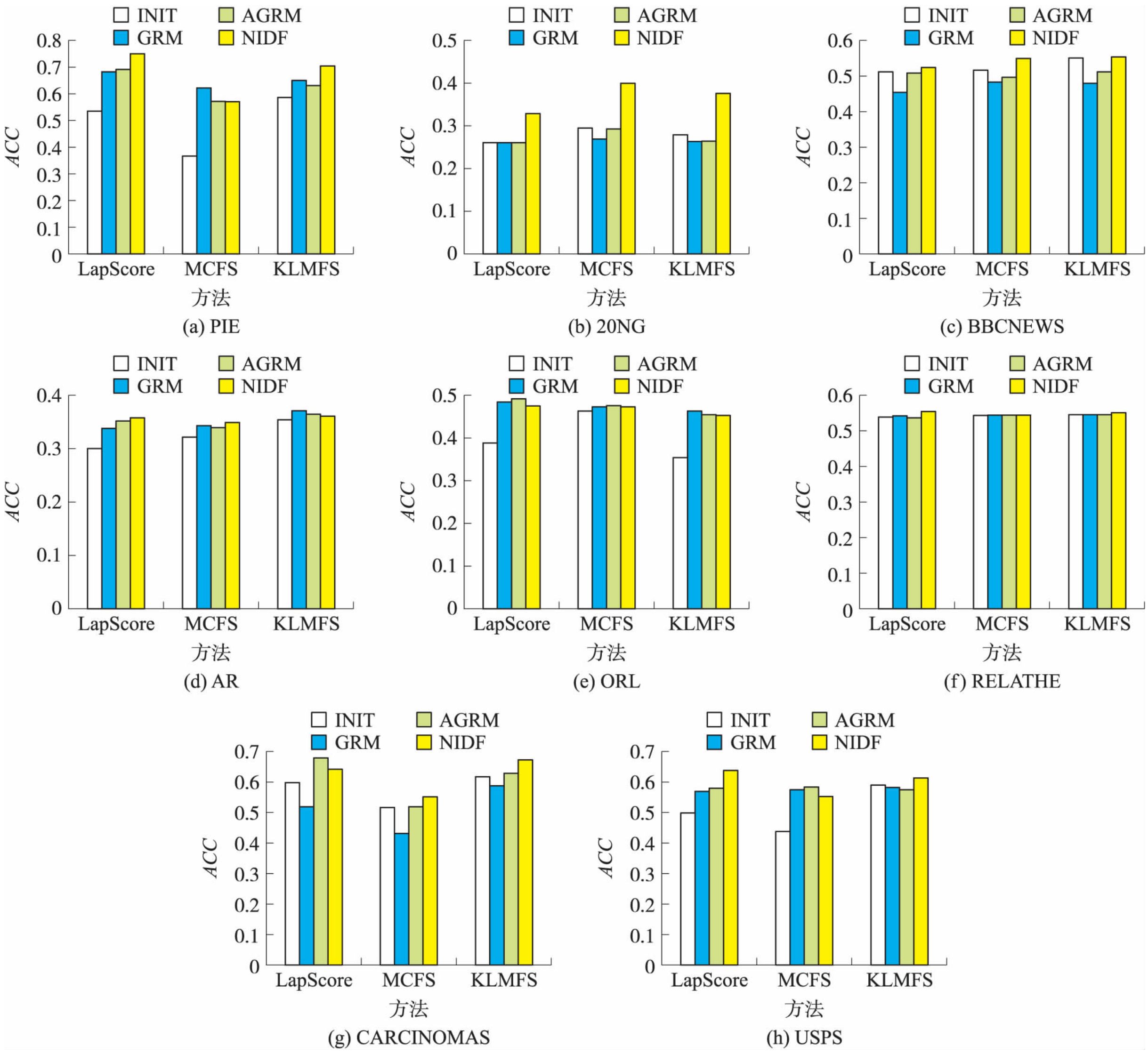

**图 4　不同数据集上 *ACC* 值的直观展示**

通过分析可以得出,经本文提出的 NIDF 模型处理后, LapScore_NIDF、MCFS_NIDF 以及 KLMFS_NIDF 相比较于原始的 LapScore、MCFS 和 KLMFS 方法,在准确性 ACC 上分别能提高将近 17.51%、15.26%和 10.50%。然而,由图 4 可以直观看出,本文提出的 NIDF 模型并不能绝对优于现有的 GRM 和 AGRM 框架,但是从总体上看,经本文提出的 NIDF 模型处理后的新方法,相比较于经 GRM 和 AGRM 处理后方法的平均值,在准确性 ACC 上分别能提高将近 7.40%、5.41% 和 8.16%。

总的来说,尽管本文提出的 NIDF 模型不能绝对优于现有的 GRM 和 AGRM 框架,但是在大多数情况下可以达到比这两个框架更好的效果,少数情况下效果相差无几。另外,本文提出的 NIDF 作为一个后处理框架,相比较于原始的无监督特征选择方法,可以达到对其性能的进一步提升,因此有一定的实用意义。

### 3.5　参数设置

在求数据集的近似区间数据集时,涉及到邻域 $k$ 和参数 $alpha$,这里设置的邻域数 $k$ 是 15,$alpha$ 是 3。由于 K-means 聚类的结果受初始值影响较大,故本文在每一次评估算法性能时重复进行 K-means聚类 20 次,最终给出平均值。本文在利用所选特征评估算法性能时,所选择的特征数集合是[10:10:100],最终取所有结果的平均数。

## 4　结束语

本文首先采用区间的方式对数据集进行近似。将一个数据集表示成与其相关几个数据集的组合,然后基于上述得到的多种数据表示,本文通过实验验证了其优劣性,并思考从全局角度对这些数据集进行处理,进而提出了一种新的模型——基于邻域区间扰动融合的无监督特征选择算法框架 NIDF。实验证明,通过对特征的最终得分和区间近似数据集的联合学习,该模型能一定程度上提高原始特征选择方法的性能,并且在大多数情况下优于现有的两个后处理框架 GRM 和 AGRM。

**参考文献:**